%% file: colm2025_conference.tex
\documentclass{article}
\usepackage[final]{colm2025_conference}

\usepackage{microtype}
\usepackage{graphicx}
\usepackage{booktabs}

\usepackage{hyperref}

\input{math_commands.tex}

\usepackage{url}

\usepackage{latexsym}
\usepackage{enumitem}
\usepackage{pifont}
\usepackage{color}
\usepackage{titletoc}
\usepackage{colortbl}
\usepackage{wrapfig}  
\usepackage{lipsum} 
\usepackage{multirow}
\usepackage{listings}
\usepackage{array}
\usepackage{booktabs}
\usepackage{caption}
\usepackage{dsfont}

\usepackage{ulem}
\usepackage{array}
\usepackage{amsfonts}
\usepackage[table]{xcolor}

\usepackage{graphicx}
\usepackage{subcaption}

\usepackage{amsmath}
\usepackage{amssymb}
\usepackage{mathtools}
\usepackage{amsthm}

\usepackage{algorithm}
\usepackage[noend]{algorithmic}

\usepackage[capitalize,noabbrev]{cleveref}

\usepackage{lineno}

\definecolor{darkblue}{rgb}{0, 0, 0.5}
\hypersetup{colorlinks=true, citecolor=darkblue, linkcolor=darkblue, urlcolor=darkblue}

\usepackage[textsize=tiny]{todonotes}

\theoremstyle{plain}
\newtheorem{theorem}{Theorem}[section]

\theoremstyle{definition}

\theoremstyle{remark}
\newtheorem{remark}[theorem]{Remark}

\newcommand{\sbb}{\mathbf{s}}

\newcommand{\xb}{\mathbf{x}}
\newcommand{\yb}{\mathbf{y}}

\newcommand{\cA}{\mathcal{A}}

\newcommand{\cD}{\mathcal{D}}

\newcommand{\cL}{\mathcal{L}}

\newcommand{\cP}{\mathcal{P}}

\newcommand{\EE}{\mathbb{E}}

\newcommand{\PP}{\mathbb{P}}

\newcommand{\btheta}{\bm{\theta}}

\DeclareMathOperator{\ind}{\mathds{1}}

\newcommand{\ours}{CITER}
\newcommand{\oursl}{\textbf{C}ollaborative \textbf{I}nference with \textbf{T}oken-l\textbf{E}vel \textbf{R}outing}

\title{CITER: Collaborative Inference for Efficient Large Language Model Decoding with Token-Level Routing}

\author{Wenhao Zheng$^1$\thanks{Equal contribution.}, Yixiao Chen$^1\footnotemark[1]$, Weitong Zhang$^1$, Souvik Kundu$^2$, Yun Li$^1$, \\
\bf{Zhengzhong Liu$^4$, Eric P. Xing$^{3,4}$, Hongyi Wang$^5$, Huaxiu Yao$^1$\thanks{Corresponding author.}} \\
$^1$The University of North Carolina at Chapel Hill, $^2$Intel, $^3$Carnegie Mellon University, \\
$^4$Mohamed bin Zayed University of Artificial Intelligence, $^5$Rutgers University\\
wenhao@cs.unc.edu, huaxiu@cs.unc.edu\\
}

\begin{document}

\ifcolmsubmission
   \linenumbers
\fi

\maketitle

\begin{abstract}
   Large language models have achieved remarkable success in various tasks but suffer from high computational costs during inference, limiting their deployment in resource-constrained applications.
   To address this issue, we propose a novel \oursl{} (\ours{}) framework that enables efficient collaboration between small and large language models (SLMs \& LLMs) through a token-level routing strategy.
   Specifically, \ours{} routes non-critical tokens to an SLM for efficiency and routes critical tokens to an LLM for generalization quality.
   We formulate router training as a policy optimization, where the router receives rewards based on both the quality of predictions and the inference costs of generation.
   This allows the router to learn to predict token-level routing scores and make routing decisions based on both the current token and the future impact of its decisions.
   To further accelerate the reward evaluation process, we introduce a shortcut which significantly reduces the costs of the reward estimation and improving the practicality of our approach.
   Extensive experiments on five benchmark datasets demonstrate that \ours{} reduces the inference costs while preserving high-quality generation, offering a promising solution for real-time and resource-constrained applications.
\end{abstract}

\section{Introduction}
\label{sec:intro}

Large language models (LLMs) have revolutionized various natural language processing tasks, from machine translation to context summarization and question answering~\citep{coleman2024llm,kamallo2024towards,eniser2024translatingrealworldcodellms}.
However, their impressive performance comes with a substantial computational costs, particularly during inference. As these models grow in size, the costs of inference becomes a significant barrier to their practical deployment, especially in real-time applications.
Therefore, there is an increasing need to reduce inference costs without compromising the quality of the generated outputs.

To address these issues, most existing approaches ~\citep{dao2022flashattentionfastmemoryefficientexact,sanh2020distilbertdistilledversionbert,kou2024cllmsconsistencylargelanguage,anagnostidis2024dynamiccontextpruningefficient} have proposed different methods to route different input queries to models of different sizes to reduce inference costs while maintaining output quality. Intuitively, small language models (SLMs) are assigned with simpler tasks demanding lower computational resources, while more complex cases are routed to LLMs to ensure response accuracy.
However, most existing works only route queries to different models once, which means that either the LLM or the SLM will handle the entire response after routing.
This one-step approach limits routing flexibility, as in many response, there is only few critical tokens need to be generated by LLM while the rest of tokens can be easily generated by SLM efficiently. As a result, simply routing these queries to LLM will significantly reduce the efficiency.

To address this challenge, we present a novel framework, namely \oursl{} (\textbf{\ours{}}). \ours{} introduces a token-level router which determines either LLM or SLM is used to generate each token. Specifically, many tokens in the response that are not important to the final prediction, can be routed and generated by SLM to reduce inference costs, while the LLM can be reserved to generate important tokens only.
We propose optimizing this router using pairwise data by reinforcement learning, with the objective of minimizing the inference costs while preserving output quality.
By employing this formulation, the router learns to predict token-level routing scores and make routing decisions not only based on the current token but also considering the impact of these decisions on future tokens.
In order to further accelerate the estimation of the reward function defined by the accuracy of the response,
we present a surrogate reward function as a shortcut, where the predictions from the SLM and LLM are leveraged to estimate the final reward without completing the whole generation process, accelerating the training process significantly.
Through this framework, we enable the collaboration of SLM and LLM
for effective and efficient autoregressive generation.

Our primary contribution is \ours{}, which reduces inference costs by employing a token-level router to select the appropriate model to generate each token. Experiments on five benchmark datasets demonstrate the effectiveness of our approach, making up to 30\% fewer inference costs with comparable accuracy or delivering up to a 25\% improvement in accuracy with the same cost compared to Co-LLM~\citep{shen2024learning}. Furthermore, our experiments in the ablation study also demonstrate that token-level routing offers more flexibility to achieve more promising results compared to query-level routing and that considering the long-term impact of routing decisions significantly boosts performance.

\noindent \textbf{Notations.} We denote $\pi_{\btheta}$ as the policy model parameterized by $\btheta$, $x_i$ as the $i$-th token in the input prompt, $y_j$ as the $j$-th token of the output response, $\ind[\cdot]$ as the indicator function and $\oplus$ as the concatenate operation. All other notations are defined prior to their first usage.

\section{\oursl{} (\ours{})}
\label{sec:method}

\begin{figure*}
   \vspace{-2em}
   \centering
   \includegraphics[width=0.9\linewidth]{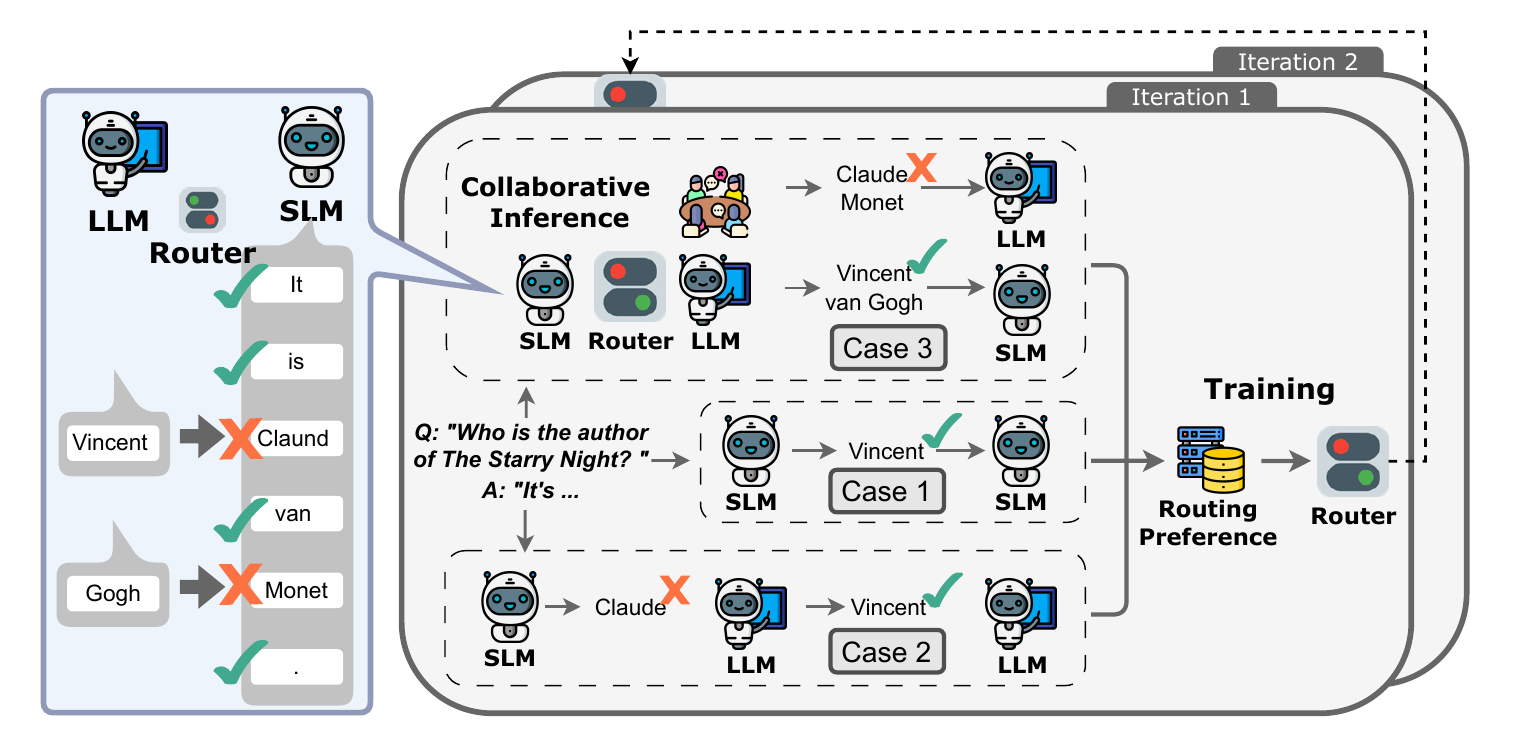}
   \caption{An overview of \ours{}. A router is leveraged to perform collaborative inference between the SLM and LLM. The router is trained using routing preference collected through three cases. \textbf{Case 1}: The SLM generates the correct token, the routing preference is assigned to the SLM. \textbf{Case 2}: The SLM generates an incorrect token, while the LLM generates the correct token, the routing preference is assigned to the LLM. \textbf{Case 3}: None of the SLM or the LLM generates the correct token, then the collaborative inference is conducted to obtain the completed response for assigning the routing preference.}
   \label{fig:main}
   \vspace{-2em}
\end{figure*}

In this section, we describe our \oursl{} (\ours{}) framework that uses token-level routing to accelerate the inference of language models. As illustrated in Figure~\ref{fig:main}, we introduce a router to facilitate collaborative inference between a powerful but computationally expensive LLM and a fast but potentially inaccurate SLM. Specifically, the router is used to predict the token-level routing score for each token, and a predefined threshold $\tau$ is used to determine which model should generate this token. The key challenge of our framework is the router training process. To feed the router with the knowledge on making the global optimal routing decisions not only based on the accuracy of the current token but also the long-term impact of its decision, we formulate the router training process as a preference-based policy optimization task, aiming to minimize the inference costs while maintaining the generation quality. To be more specific, we first formulate the RL problem and derive the reward function as token-wise routing preference, which should be computed to collect during the router training process. Subsequently, we introduce a shortcut for the reward function estimation, leveraging both the SLM and LLM's prediction to estimate the reward, significantly accelerating the collection process of the token-wise routing preference. Finally, we propose an iterative router training process to mitigate the potential inconsistencies of routing decisions in the preference collection phase and deployment. In the rest of this section, we will outline router training and collaborative inference processes in detail.

\vspace{-1em}
\subsection{Reinforcement Learning for Router Optimization}
\label{subsec:router}
We start by introducing the foundational concepts and notation for the Markov Decision Process for token-level routing. In particular, we formulate the token-level routing task as a Markov decision process (MDP)~\citep{bellman1957dynamic} where \emph{state} is a series of tokens $\sbb_{h} = (x_0, \cdots x_m, y_0, \cdots, y_h)$, including both the input prompt $(x_0, \cdots x_m)$ and the response $(y_0, \cdots, y_h)$. At each time step $h$, the agent selects its \emph{action} from $\cA = \{A_L, A_S\}$, which means generating a token from LLM ($A_L$) or SLM ($A_S$), respectively. Then we write the generation of the next token by the following \emph{transition kernel} $\PP(\sbb_{h+1} | \sbb_{h}, a_h)$ given by the dynamics of LLM and SLM. This generation process ends once the terminal token \texttt{<EOS>} is generated from either of these models. The generated token length is denoted as $H$, which can be flexible. The \emph{reward} $r(\sbb_h, a_h)$ is then related to the generation cost and the accuracy of the final response $r(\sbb_H)$. The state-action value function is defined by
\begin{align}
   \textstyle{Q_h^\pi(\sbb, a) = \EE\left[\sum_{t=h}^H r(\sbb_t, a_t) \middle | \sbb_h = \sbb, a_h = a, \pi\right]}, \label{eq:def-q}
\end{align}
with the optimal state-action value function $Q^*$ defined as $Q^*_h(\sbb, a) = \max_{\pi} Q^{\pi}_h(\sbb, a)$. The objective of the routing policy can be written by
\begin{align}
   \pi_h^*(a | \sbb) = \argmax_{\pi} \EE \left[Q_h^*(\sbb, a) - \beta \mathrm{KL}(\pi \parallel \mu)\right] \propto \mu(a | \sbb) \exp(\beta^{-1}Q_h^*(a | \sbb)),
\end{align}
where $\mu$ is the prior policy related to the cost difference for evaluating LLM or SLM. The expectation is taken over the randomness of language models, policy $\pi$ and the prompt $\sbb_0$.

\vspace{-1em}
\subsection{Preference-based Token-level Policy Optimization}
\vspace{-0.5em}

\label{subsubsec:collection}
Generally, defining the reward $r(\sbb_h, a_h)$ as well as the state-action value function $Q_h(\sbb_h, a_h)$ is difficult and may result in reward hacking~\citep{amodei2016concrete,specificationgaming2020}.
To tackle with this issue, similar with~\citep{rafailov2023direct}, we inject the pairwise preference $\ind_h[a_S \succ a_L]$ following the Bradley–Terry model~\citep{bradley1952rank} as:
\begin{align}
   \Pr_h(a_S \succ a_L | \sbb_h) = \sigma(\beta(Q_h^*(\sbb_h, a_S) - Q_h^*(\sbb_h, a_L))), \label{eq:bt}
\end{align}
where $\sigma(z)$ is the sigmoid function. Following~\citet{rafailov2024direct}, we have
\begin{align}
   Q_h^*(\sbb_h, a_L) - Q_h^*(\sbb_h, a_S)
   = & \beta \log \frac{\pi_h^*(a_L | \sbb_h)}{\mu(a_L)} - \beta \log \frac{\pi_h^*(a_S | \sbb_h)}{\mu(a_S)} \notag          \\
   = & \beta \log \frac{\pi_h^*(a_L | \sbb_h)}{\pi_h^*(a_S | \sbb_h)} - \beta \log \frac{\mu(a_L)}{\mu(a_S)}, \label{eq:dpo}
\end{align}
In the case of $\mu(a_L) = \mu(a_S)$ and $\beta = 1$,
plugging~\eqref{eq:dpo} into~\eqref{eq:bt} yields
\begin{align}
   \Pr(a_S \succ a_L | \sbb_h) & = \frac{1}{1 + \frac{\pi_h^*(a_L | \sbb_h)}{\pi_h^*(a_S | \sbb_h)}} = \pi_h^*(a_S | \sbb_h), \notag
\end{align}
where the latter equation is due to the fact that $\pi_h^*(a_S | \sbb_h) + \pi_h^*(a_L | \sbb_h) = 1$. Therefore, given a sequence of token $\sbb_h$, once we have labeled the preference $\ind[a_S \succ a_L | \sbb_h]$, $\pi_h(a_S | \sbb_h)$, the routing agent $\pi$ can be learned by minimizing the cross-entropy loss
\begin{align}
   \label{eq:loss}
   \cL(\btheta)  = -\sum_{\sbb_h} ( \ind_h[a_S \succ a_L | \sbb_h] \log \pi_h(a_S | \sbb_h, \btheta)
   + \ind_h[a_L \succ a_S | \sbb_h] \log \pi_h(a_L | \sbb_h, \btheta) ),
\end{align}
where $\ind[a_L \succ a_S | \sbb_h]$ indicates using large language model is preferred at state $\sbb_h$.

\vspace{-1em}
\subsection{Acquiring Token-level Routing Preference} \label{subsec:acq}
\vspace{-0.5em}

In this subsection, we describe our strategy to determine the preference label $\ind[a_L \succ a_S | \sbb_h]$. For a state $\sbb_h$, we first generate the next token $y_{h+1}$ with the small language model and then complete the whole trajectory $\sbb_{H}$ until the generation ends with $\texttt{<EOS>}$ using the routing policy $\pi$. Compared to~\eqref{eq:def-q}, the reward collected on this trajectory $\sbb_H$ is an unbiased estimation of $Q_h^{\pi}(\sbb_h, a_S)$. Intuitively, if using the small language model in the current step $h$ can generate the correct response $\sbb_H$, then the small language model is preferred. Otherwise, we assign $a_L \succ a_S$ and assume that the large language model can generate the correct answer, as implemented in Line~\ref{ln:shortcut-2} in Algorithm~\ref{alg:main}.

However, generating and evaluating the final response $\sbb_H$ might be expensive or even infeasible when $H$ is large. In order to further accelerate the token-level routing preference label, we introduce a \emph{shortcut} by leveraging the ground truth response $\sbb^*_H$ provided in the dataset. As Line~\ref{ln:shortcut-1} in Algorithm~\ref{alg:main} suggests, if the next token $y_{h+1}^S$ generated by the small language model is exactly the same as the ground-truth token $y^*_{h+1}$, we assign $a_S \succ a_L$ since the behavior of the small language model is good enough to match the ground-truth model. In the case where the next token generated by the small language model does not match the ground truth, as carried out in Line~\ref{ln:default}, we check the next token generated by the large language model $y_{h+1}^L$ and assign $a_L \succ a_S$ if $y_{h+1}^L = y_{h+1}^*$. Otherwise we will go back to the aforementioned case to evaluate the completed generated trajectory as conducted in Line~\ref{ln:default} in Algorithm~\ref{alg:main}.
We would like to highlight that only when both models fail to generate the correct token $y^*_h$ based on ground truth context, the full response generation is required to compute the reward. This shortcut allows us to obtain routing preferences for most tokens without generating the full response. Empirically, we find that about $80\% \sim 90\%$ of the tokens can be correctly predicted by either the SLM or LLM, which makes the shortcut significantly reduce the inference costs of estimating the reward function.

\begin{algorithm}[t]
   \caption{Preference-based Router Optimization for \ours{}}
   \label{alg:main}
   \begin{algorithmic}[1]
      \STATE \textbf{Input:} Training data $\cD = \{\xb, \yb^*\}$, SLM and LLM $M_S, M_L$, number of rounds $K$.
      \STATE Initialize training policy $\pi_{\btheta}$, preference set $\cP_0 = \emptyset$
      \FOR{round $k = 1, \cdots, K$}
      \STATE Initialize preference set $\cP_k = \emptyset$
      \FOR{prompt-response pair $\xb, \yb^*$ \textbf{in} $\cD$}
      \STATE Set $h = 0$, $\sbb_0 = \xb$
      \WHILE{$y_h$ is not \texttt{<EOS>}}
      \STATE \textbf{if } $M_S(\sbb_h) = y^*_{h+1}$ \label{ln:shortcut-1} \textbf{then} Set $p_h = 1$ {\color{blue} \texttt{/* Case 1. }$a_S \succ a_L$\texttt{ */}}
      \STATE \textbf{if } $M_L(\sbb_h) = y^*_{h+1}$ \label{ln:shortcut-2} \textbf{then} Set $p_h = 0$ {\color{blue} \texttt{/* Case 2. }$a_L \succ a_S$\texttt{ */}}
      \STATE \textbf{else } {\color{blue} \texttt{/* Case 3. */}} \label{ln:default}
      \STATE Generate new token: $\tilde \sbb_{h+1} = \sbb_h \oplus [M_S(\sbb_h)]$
      \STATE Generate \begin{small}$\tilde \sbb = \textbf{\ours{}}(\tilde \sbb_{h+1}, M_S, M_L, \pi_{\btheta}, \frac12)$\end{small}
      \STATE $p_h = 1$ \textbf{if } $\tilde \sbb$ is \texttt{correct} \textbf{else} $p_h = 0$.
      \STATE Update $\cP_k = \cP_k \cup \{\sbb_h, p_h\}$
      \ENDWHILE
      \ENDFOR
      \STATE {\color{blue} \texttt{/* Preference-based Optimization */}}
      \STATE Update $\btheta$ by minimizing loss
      $\cL(\btheta) = -\sum_{(\sbb, p) \in \cP_k}p\log \pi_{\btheta}(a_S | \sbb) + (1 - p) \log \pi_{\btheta}(a_L | \sbb)$
      \ENDFOR
      \STATE \textbf{Output:} Routing policy $\pi_{\btheta}$.
   \end{algorithmic}
\end{algorithm}

\begin{algorithm}[t]
   \caption{\oursl{}~(\textbf{\ours{}})}
   \label{alg:inf}
   \begin{algorithmic}[1]
      \STATE \textbf{Input:} Input prompt $\sbb$, SLM and LLM $M_S, M_L$, policy $\pi_{\btheta}$, threshold $\tau$
      \STATE Let $\tilde \sbb = \sbb$
      \WHILE {\textbf{True}}
      \STATE Set $M = M_S$ \textbf{if} $\pi(a_S | \tilde \sbb) \ge \tau$ \textbf{else} $M = M_L$
      \STATE Generate next token and let $\tilde \sbb = \sbb \oplus \{M(\tilde \sbb)\}$
      \STATE \textbf{if} $M(\tilde \sbb) = \texttt{<EOS>}$ \textbf{then break}
      \ENDWHILE
      \STATE \textbf{Output:} Generated response $\tilde \sbb$.
   \end{algorithmic}
\end{algorithm}

\vspace{-1em}
\subsection{Proposed Algorithm}
\vspace{-0.5em}

Finally we summarize the proposed algorithm as well as some implementation details in Algorithm~\ref{alg:main} as an iterative update of the routing policy $\pi_{\btheta}$. In each iteration $k$, the router $\pi_{\btheta_{k-1}}$ from the previous iteration is used to collect routing preferences $\cP = \{\sbb, p\}$. Then iteration goes for at most $K$ rounds but can also stop early when $\cP_k = \cP_{k-1}$ and thus the policy optimization converges. The preference $p \in \{0, 1\}$ is labeled through the three cases described in Subsection~\ref{subsec:acq}, where the algorithm calls the inference algorithm~\textbf{\ours{}} when neither the LLM nor the SLM can predict the correct token.

The inference algorithm~\textbf{\ours{}} is presented in Algorithm~\ref{alg:inf}. In detail, \textbf{\ours{}} uses a deterministic policy where it chooses the small language model when $\pi(a_S | \sbb) \ge \pi(a_L | \sbb)$ (i.e., $\pi(a_S | \sbb) \ge \frac12$) and vice versa. To further investigate the balance between efficiency and precision by collaborating with LLM and SLM, we introduce another layer of prior policy $(\rho(a_S), \rho(a_L))$, where $\rho(a_S) + \rho(a_L) = 1$. Thus, the deterministic rule of selecting the SLM from the posterior policy distribution $\pi'(a | \sbb) \propto \pi(a | \sbb)\rho(a)$ is that
\begin{align}
   \pi(a_S | \sbb) \rho(a_S) \ge \pi(a_L | \sbb) \rho(a_L) \Rightarrow \pi(a_S | \sbb) \ge \rho(a_L), \notag
\end{align}
where we introduce $\tau := \rho(a_L)$ as a hyper parameter in the algorithm to probe this balance.

\begin{remark}
   We maintain separate KV caches for SLM and LLM. When CITER switches between them, the previous KV cache is preserved, allowing it to be reused when switching back. This eliminates the need for redundant computations, improving efficiency.
\end{remark}

\vspace{-1em}
\section{Experiments}
\label{sec:experiments}
\vspace{-1em}

In this section, we evaluate the performance of \ours{} aiming to answer the following questions: (1) Compared with the previous works on speeding up the inference of LLM, how does our framework perform in terms of the inference costs and the response quality? (2) Does the components we proposed in our framework boost the performance of the router? (3) Does the iterative training process of the router improve the performance of our framework? (4) How does the performance of our framework change with the size of the LLM? (5) Can the router distinguish the critical and non-critical tokens correctly?

\vspace{-1em}
\subsection{Experimental Setup}
\label{subsec:exprimental}
\vspace{-0.5em}

\begin{table*}[t]
   \centering
   \small
   \vspace{-1em}
   \caption{The statistics of our evaluation datasets. The commonsense QA dataset and MMLU-Professional Psychology dataset are denoted as CS QA and MMLU-PP, respectively.}
   \label{tab:dataset}
   \resizebox{0.95\linewidth}{!}{
       \begin{tabular}{l|ccccc}
          \toprule
          \textbf{Dataset}       & \textbf{Domain} & \textbf{Task}      & \textbf{\# choices} & \textbf{Train size} & \textbf{Test size} \\
          \midrule
          Commonsense QA (CS QA) & General         & Multi-choice       & 5                   & 9,741               & 1,221              \\
          ARC-Challenge          & Reasoning       & Multi-choice       & 4                   & 1,119               & 299                \\
          MMLU-PP                & Psychology      & Multi-choice       & 4                   & 612                 & 69                 \\
          GSM8k                  & Math            & Question answering & N/A                 & 7,473               & 1,319              \\
          MATH                   & Math            & Question answering & N/A                 & 7,500               & 5,000              \\
          \bottomrule
       \end{tabular}
   }
   \vspace{-1em}
\end{table*}

\begin{figure*}[t]
   \centering
   \begin{subfigure}{0.195\linewidth}
      \centering
      \includegraphics[width=\linewidth]{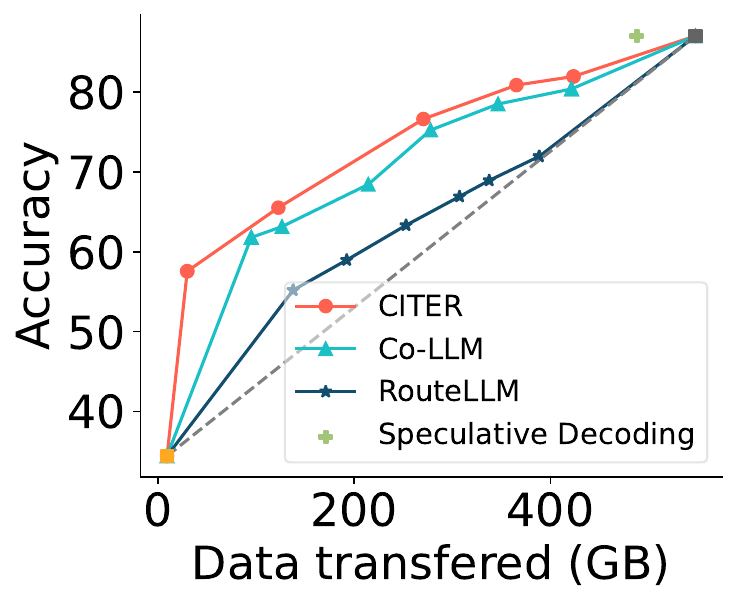}
      \caption{CS QA}
   \end{subfigure}
   \centering
   \centering
   \begin{subfigure}{0.195\linewidth}
      \centering
      \includegraphics[width=\linewidth]{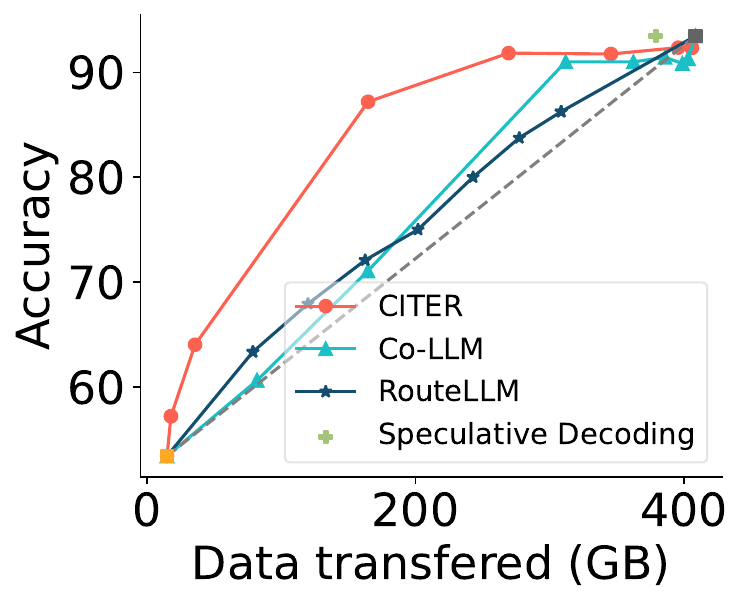}
      \caption{ARC-Challenge}
   \end{subfigure}
   \centering
   \begin{subfigure}{0.195\linewidth}
      \centering
      \includegraphics[width=\linewidth]{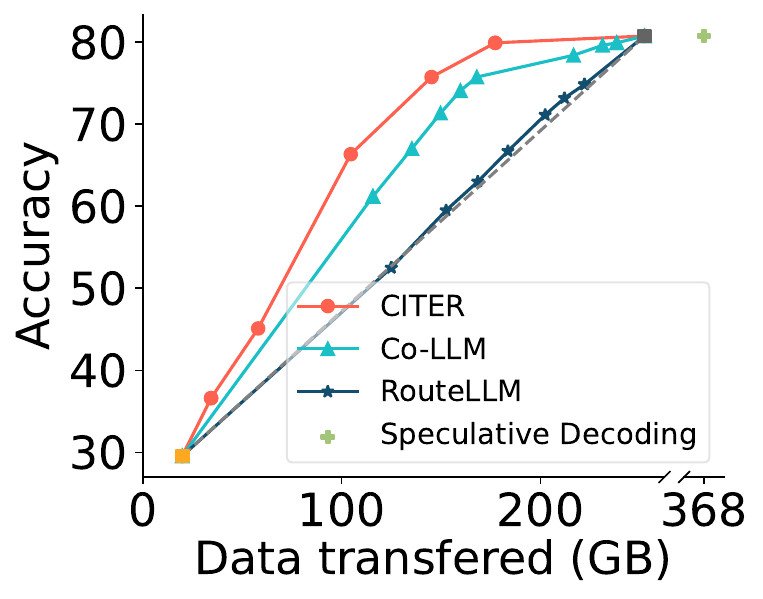}
      \caption{GSM8k}
   \end{subfigure}
   \centering
   \begin{subfigure}{0.195\linewidth}
      \centering
      \includegraphics[width=\linewidth]{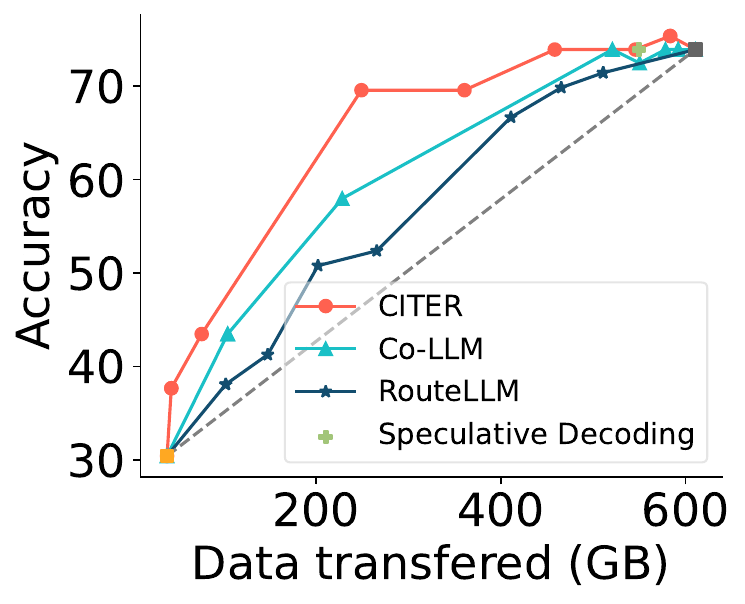}
      \caption{MMLU-PP}
   \end{subfigure}
   \centering
   \begin{subfigure}{0.195\linewidth}
      \centering
      \includegraphics[width=\linewidth]{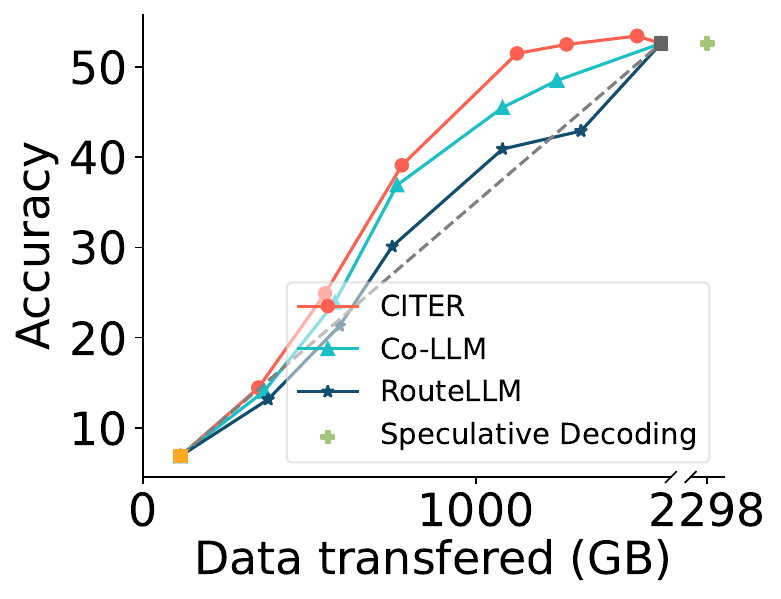}
      \caption{MATH}
   \end{subfigure}
   \caption{The accuracy vs data transformation amount curve of \ours{} and the baselines. The yellow and grey squares represent the performance of slm and llm respectively. The grey line represents the random routing strategy. Points closer to the top-left corner indicate better acceleration performance.}
   \label{fig:overall}
   \vspace{-1em}
\end{figure*}

\noindent \textbf{Dataset Description.}
\label{para:dataset}
We evaluate \ours{} and our baselines on five widely-used academic benchmark datasets: the commonsense QA dataset~\citep{talmor2019commonsenseqa} contains 12,102 questions requiring different types of commonsense knowledge to answer; the ARC-Challenge dataset~\citep{peter2018arc}, including 1,418 genuine grade-school level, multiple-choice science questions; the MMLU-Professional Psychology dataset~\citep{hendryckstest2021mmlu}, consisting of 874 multiple-choice questions on psychology; the GSM8k dataset~\citep{cobbe2021gsm8k} with 8.5K high quality linguistically diverse grade school math word problems and MATH dataset~\citep{hendrycksmath2021} with 12.5k problems from mathematics competitions respectively. The statistics of the datasets are in Table~\ref{tab:dataset}.

\noindent \textbf{Evaluation.}
\label{para:evaluation}
To evaluate the performance of \ours{} and the baseline methods, we use the accuracy of responses to reflect response quality and define the inference cost as the average amount of data transformation per sample, mainly the KV cache that must be transferred from GPU HBM to the on-chip cache, since LLM generation is primarily memory-bound.
Details on the data transformation calculations and an illustration of the memory-bound nature of LLM generation are provided in Appendix~\ref{app:costs}.
Additionally, for both query-level routing methods and token-level routing methods, a threshold $\tau$ is applied in each method to balance the trade-off between leveraging the LLM to improve the response quality or offloading to the SLM to reduce the overall inference costs.
We then plot the accuracy curve versus the average amount of data transformation per sample to illustrate the acceleration performance of both \ours{} and the baselines. The optimal point is located in the top-left corner of the curve, corresponding to the highest accuracy with the lowest costs.

\noindent \textbf{Baselines.}
\label{para:baselines}
We compare \ours{} against three methods: a representative query-level routing approach (RouteLLM~\citep{ong2024routellm}), a token-level routing method (Co-LLM~\citep{shen2024learning}), and a non-routing-based technique (Speculative Decoding~\citep{leviathan2023fastinferencetransformersspeculative}). RouteLLM makes routing decisions at the query level, directing entire queries to different models for generation, while Co-LLM operates at the token level, dynamically routing each token to different models throughout the generation process. In contrast, Speculative Decoding does not involve routing between models; instead, it leverages the SLM to propose a set of candidate tokens, and then verify them by the LLM.

\vspace{-0.5em}
\noindent \textbf{Implementation Details}
\label{para:implementation}
We implement our framework using the Hugging Face Transformers library~\citep{wolf2020transformers}. For the SLM and LLM, we utilize Qwen2-1.5b and Qwen2-72b, respectively. The router is implemented as a multilayer perceptron (MLP) network with three hidden layers, ReLU activation~\citep{agarap2019deeplearningusingrectified}, BatchNorm normalization~\citep{ioffe2015batchnormalizationacceleratingdeep}, and a $0.1$ dropout rate. It is trained using the Adam optimizer~\citep{kingma2017adammethodstochasticoptimization} with a learning rate of $1 \times 10^{-7}$, betas of $(0.9, 0.99)$, and no weight decay. Training is performed on a single NVIDIA H100 GPU with a batch size of $80$. The iterative training process runs for $2$ rounds. We use the hidden state corresponding to the last generated token from the SLM as the input to our router. This approach enables the router to utilize the rich representations extracted by the SLM, allowing routing decisions to be informed not only by the current token but also by the broader context accumulated thus far.

\begin{figure*}[t]
   \centering
   \begin{subfigure}{0.195\linewidth}
      \centering
      \includegraphics[width=\linewidth]{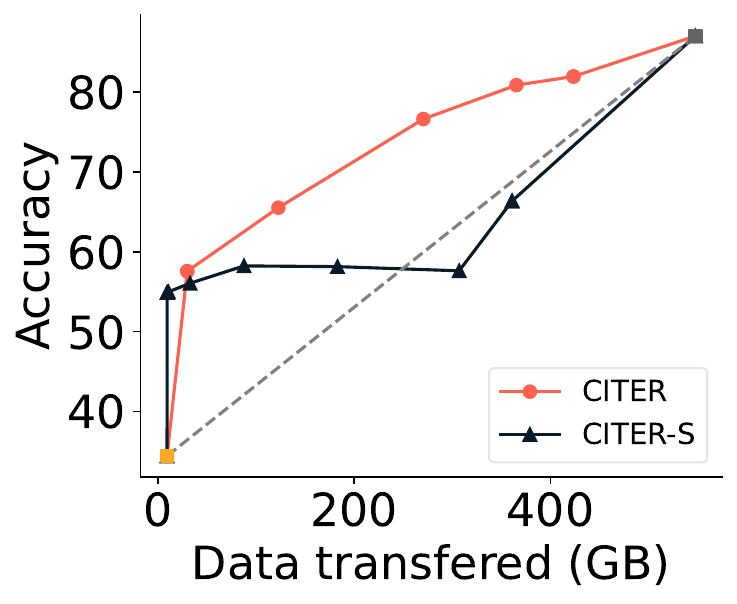}
      \caption{CS QA}
   \end{subfigure}
   \centering
   \begin{subfigure}{0.195\linewidth}
      \centering
      \includegraphics[width=\linewidth]{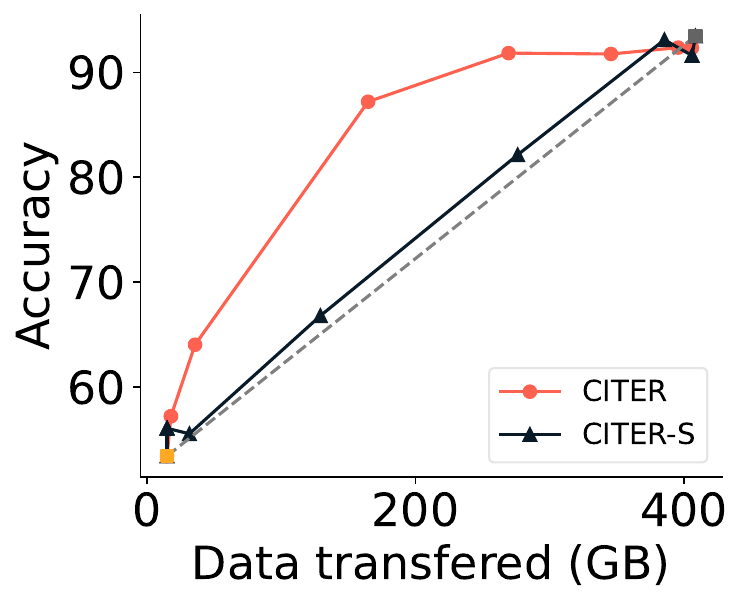}
      \caption{ARC-Challenge}
   \end{subfigure}
   \centering
   \begin{subfigure}{0.195\linewidth}
      \centering
      \includegraphics[width=\linewidth]{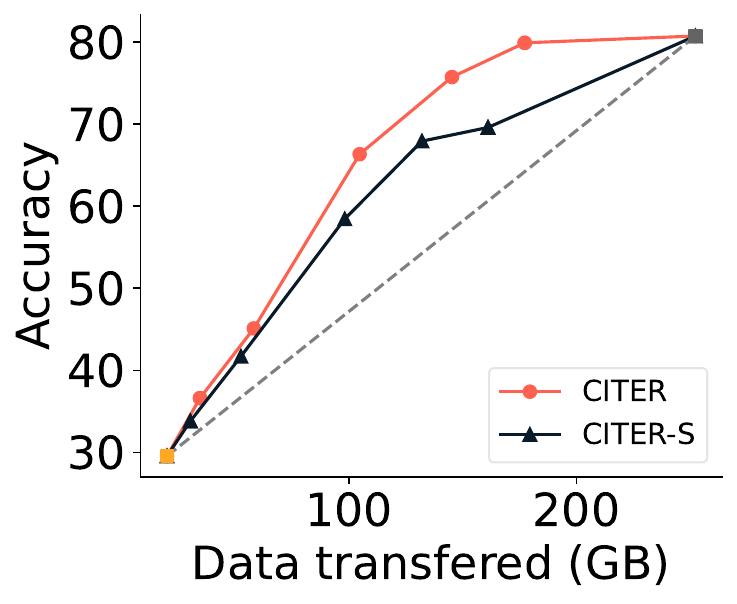}
      \caption{GSM8k}
   \end{subfigure}
   \centering
   \begin{subfigure}{0.195\linewidth}
      \centering
      \includegraphics[width=\linewidth]{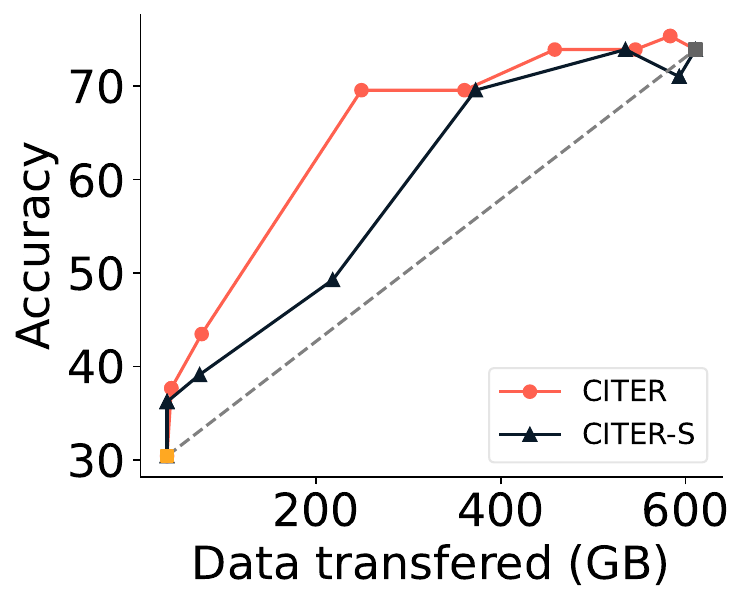}
      \caption{MMLU-PP}
   \end{subfigure}
   \begin{subfigure}{0.195\linewidth}
      \centering
      \includegraphics[width=\linewidth]{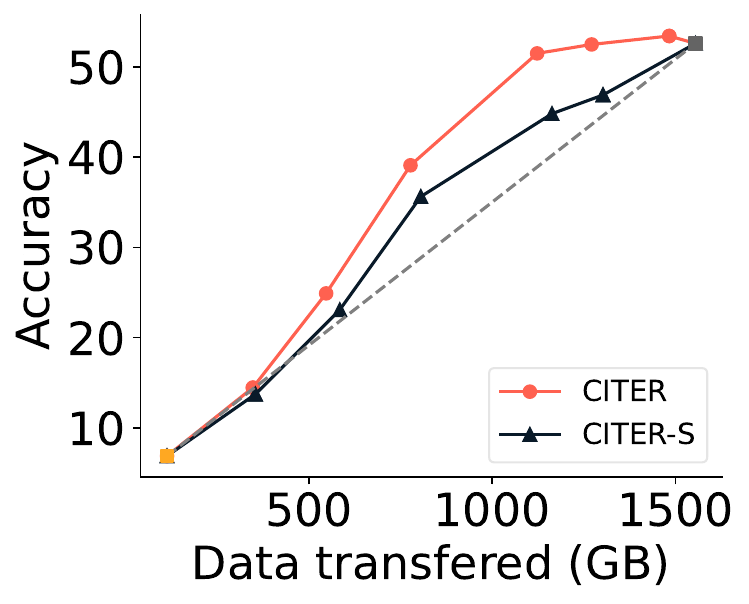}
      \caption{MATH}
   \end{subfigure}
   \caption{The accuracy vs data transformation amount curve of \ours{} and the varient \ours{}-S. The yellow and grey squares represent the performance of slm and llm respectively. The grey line represents the random routing strategy. Points closer to the top-left corner indicate better acceleration performance.}
   \label{fig:ablation}
   \vspace{-2em}
\end{figure*}

\vspace{-1em}
\subsection{Overall Performance}
\label{subsec:overall}
\vspace{-0.5em}

We conduct extensive experiments to assess the performance of \ours{} across all benchmark datasets, comparing it against baseline methods. The results are presented in Figure~\ref{fig:overall}. Clearly, all token-level routing methods, including \ours{} and Co-LLM, significantly outperform the query-level routing method, RouteLLM, across all datasets, particularly on the Commonsense QA and GSM8k datasets, reducing up to 30\% inference costs while maintaining the same accuracy or achieving up to 12\% higher accuracy with the same cost. This emphasizes the effectiveness of token-level routing, which provides enhanced flexibility in reducing inference costs while preserving response quality. Notably, Speculative Decoding does reduce inference costs on some multiple-choice datasets. However, its verification mechanism requires the small model to produce outputs identical to those of the large model to maintain lossless output quality, which is overly stringent and limits the potential for further cost reduction on complex cases. As a result, on mathematical datasets, the acceptance rate of candidate tokens proposed by the small model is extremely low, leading to higher inference costs than simply using the large model alone, which is unacceptable. Furthermore, \ours{} consistently surpasses Co-LLM, achieving comparable accuracy with up to 27\% fewer inference costs or delivering up to a 17\% improvement in accuracy with the same cost. These findings demonstrate the success of our framework in accelerating LLM inference. This outcome is expected, as Co-LLM does not consider long-term information during the router training phase, which is crucial for token-level routing. In the following section, we present experiments to further demonstrate the importance of incorporating long-term information in router training.

\begin{figure*}[t]
   \centering
   \begin{subfigure}{0.195\linewidth}
      \centering
      \includegraphics[width=\linewidth]{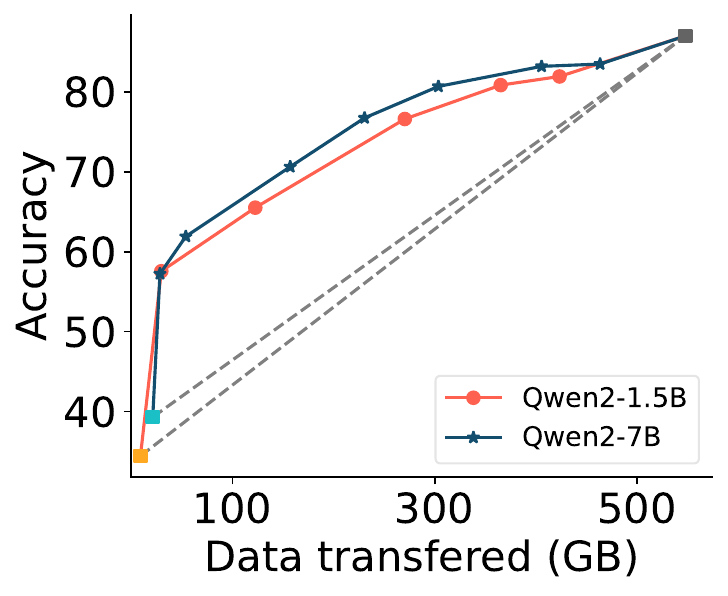}
      \caption{CS QA}
   \end{subfigure}
   \centering
   \begin{subfigure}{0.195\linewidth}
      \centering
      \includegraphics[width=\linewidth]{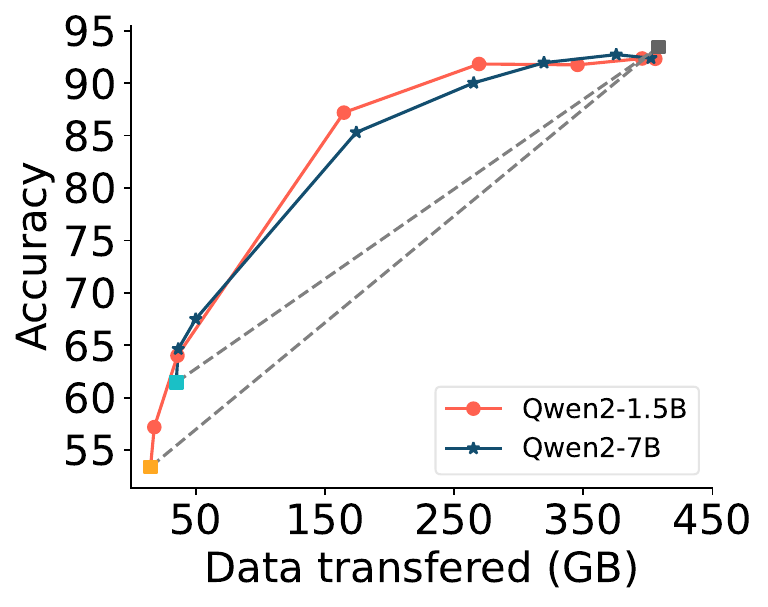}
      \caption{ARC-Challenge}
   \end{subfigure}
   \centering
   \begin{subfigure}{0.195\linewidth}
      \centering
      \includegraphics[width=\linewidth]{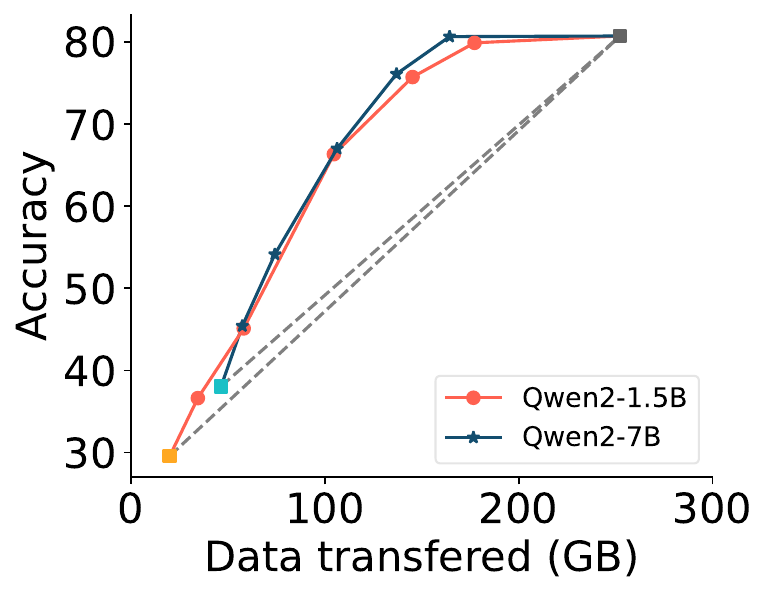}
      \caption{GSM8k}
   \end{subfigure}
   \centering
   \begin{subfigure}{0.195\linewidth}
      \centering
      \includegraphics[width=\linewidth]{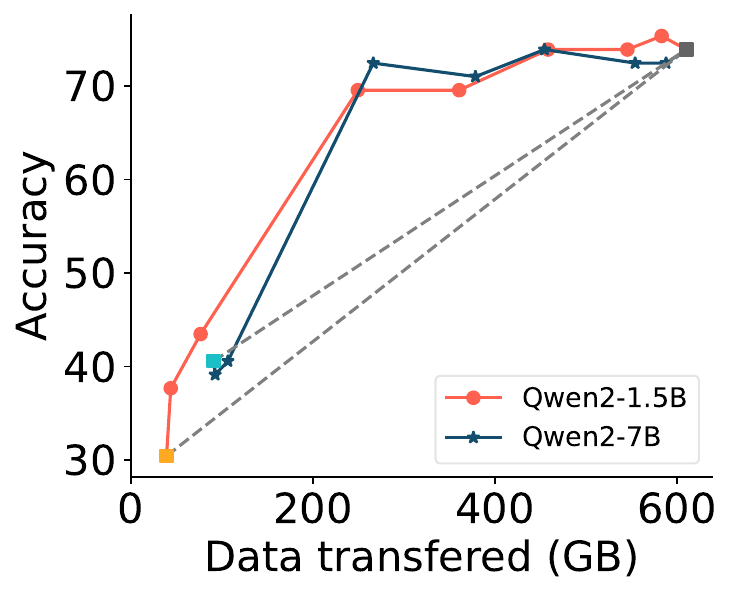}
      \caption{MMLU-PP}
   \end{subfigure}
   \centering
   \begin{subfigure}{0.195\linewidth}
      \centering
      \includegraphics[width=\linewidth]{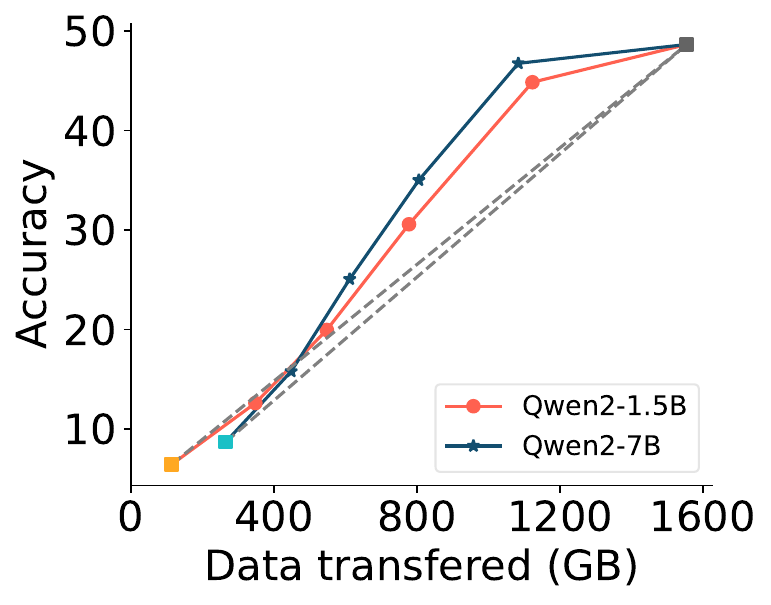}
      \caption{MATH}
   \end{subfigure}
   \caption{The accuracy vs data transformation amount curve of \ours{} with 1.5B SLM and \ours{} with 7B SLM. The yellow, blue and grey squares represent the performance of Qwen2-1.5B, Qwen2-7B and Qwen2-72B respectively. The grey line represents the random routing strategy. Points closer to the top-left corner indicate better acceleration performance.}
   \label{fig:7b}
   \vspace{-1em}
\end{figure*}

\begin{wrapfigure}{r}{0.3\linewidth}
   \vspace{-5em}
   \includegraphics[width=\linewidth]{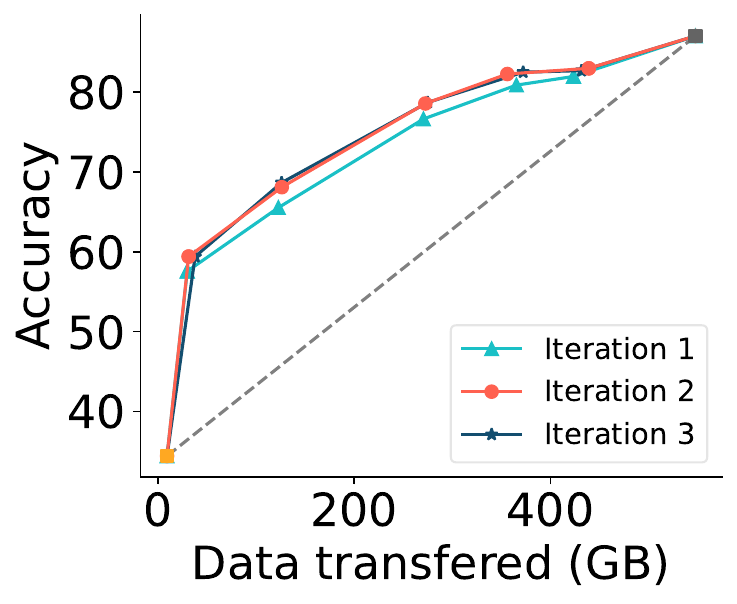}
   \caption{Accuracy vs. inference costs of \ours{} with router over the first three iterations on the commonsense QA datasets. Points closer to the top-left corner indicate better acceleration performance.}
   \label{fig:iterative}
   \vspace{-3em}
\end{wrapfigure}

\vspace{-1em}
\subsection{Analysis of Long-Term Influence}
\label{subsec:ablation}
\vspace{-0.5em}

In this section, we conduct an ablation study on a key component of our framework: the long-term influence of routing decisions, to evaluate its effectiveness. For this purpose, we design an ablation variant, \ours{}-S, where the SLM is selected if both the SLM and LLM provide incorrect predictions during the routing preference collection, disregarding the long-term impact of routing decisions. The results are shown in Figure~\ref{fig:ablation}. Clearly, \ours{} significantly outperforms the ablation variant \ours{}-S across all datasets, reducing inference costs by up to 42\% while maintaining the same accuracy, or achieving up to a 23\% accuracy improvement with the same cost. These findings highlight the critical role of accounting for the long-term influence of routing decisions.

\subsection{Analysis of Iterative Training Process}
\label{subsec:iterative}

To highlight the importance of the iterative training process, we present the performance curve of \ours{} with the router over the first three iterations on the Commonsense QA dataset. As shown in Figure~\ref{fig:iterative}, the results demonstrate a clear improvement in performance in the first two iterations. In the second iteration, \ours{} reduces $\sim 5\%$ inference costs while maintaining the same accuracy or achieves $2 \sim 3\%$ higher accuracy with the same cost compared to the first. This improvement underscores the effectiveness of our proposed iterative training process. Moreover, the performance curve of the third iteration closely follows that of the second, indicating that the router has already converged by the second iteration. The rapid convergence of the router emphasizes the robustness of our training strategy, suggesting that optimal performance can be achieved without excessive costs.

\vspace{-1em}
\subsection{Results on Different Model Families}
\label{subsec:modelbrand}
\vspace{-0.5em}

\begin{figure*}[t]
   \centering
   \begin{subfigure}{0.195\linewidth}
      \centering
      \includegraphics[width=\linewidth]{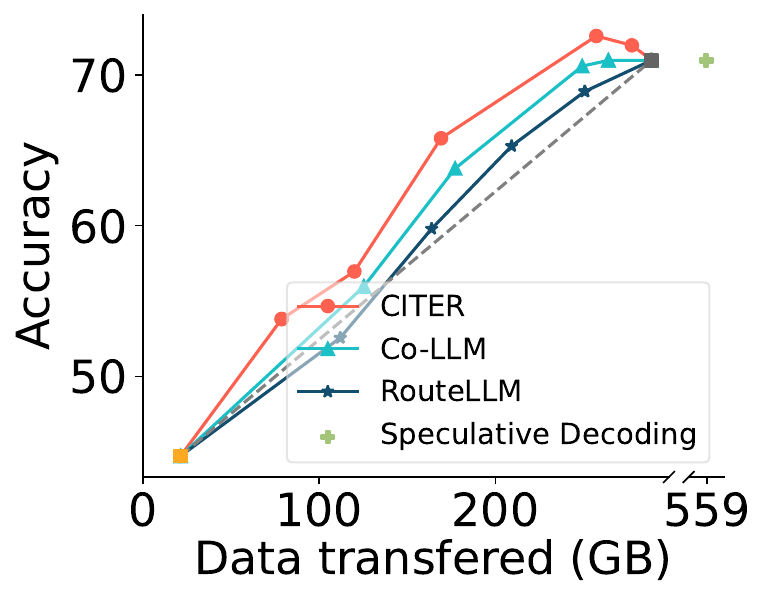}
      \caption{CS QA}
   \end{subfigure}
   \centering
   \centering
   \begin{subfigure}{0.195\linewidth}
      \centering
      \includegraphics[width=\linewidth]{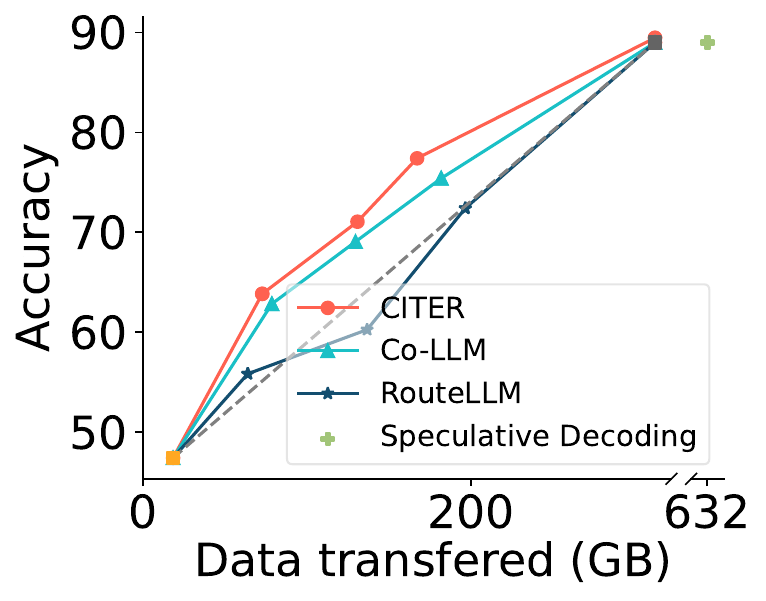}
      \caption{ARC-Challenge}
   \end{subfigure}
   \centering
   \begin{subfigure}{0.195\linewidth}
      \centering
      \includegraphics[width=\linewidth]{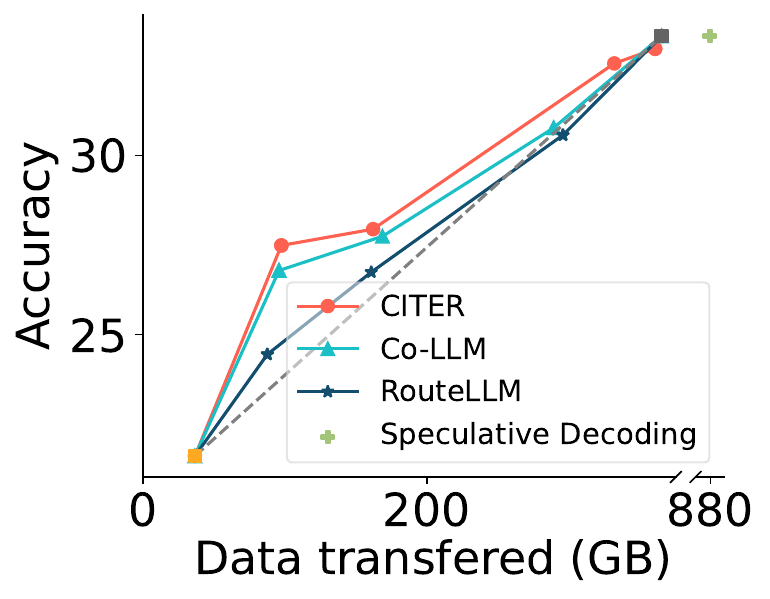}
      \caption{GSM8k}
   \end{subfigure}
   \centering
   \begin{subfigure}{0.195\linewidth}
      \centering
      \includegraphics[width=\linewidth]{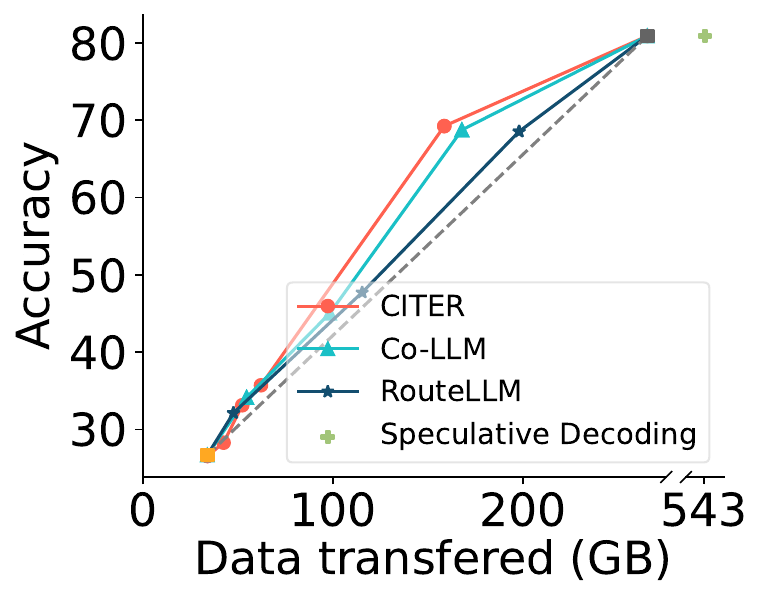}
      \caption{MMLU-PP}
   \end{subfigure}
   \centering
   \begin{subfigure}{0.195\linewidth}
      \centering
      \includegraphics[width=\linewidth]{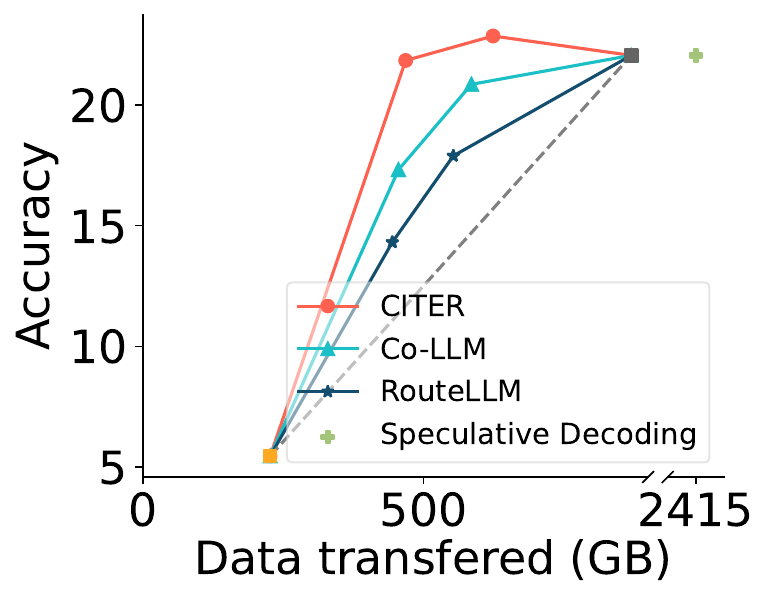}
      \caption{MATH}
   \end{subfigure}
   \caption{The accuracy vs data transformation amount curve of \ours{} and the baselines with Llama3.1 series. The yellow and grey squares represent the performance of slm and llm respectively. The grey line represents the random routing strategy. Points closer to the top-left corner indicate better acceleration performance.}
   \label{fig:llama}
   \vspace{-1em}
\end{figure*}

Additionally, we conduct experiments with Llama3.1 series models to demonstrate the compatibility of our framework. Specifically, we leverage the Llama3.1-70B model as the LLM and the Llama3.1-8B model as the SLM. The results are illustrated in Figure~\ref{fig:llama}. Similarly to the results with Qwen series, \ours{} consistently outperforms all other baseline methods, achieving comparable accuracy with up to 32\% fewer inference costs or providing up to a 5\% improvement in accuracy with the same cost, compared to Co-LLM, the best baseline method. This result further demonstrates the effectiveness of our framework and additionally highlights the compatibility of our framework with different series of models.

\begin{wrapfigure}{r}{0.3\linewidth}
   \vspace{-3em}
   \centering
   \includegraphics[width=\linewidth]{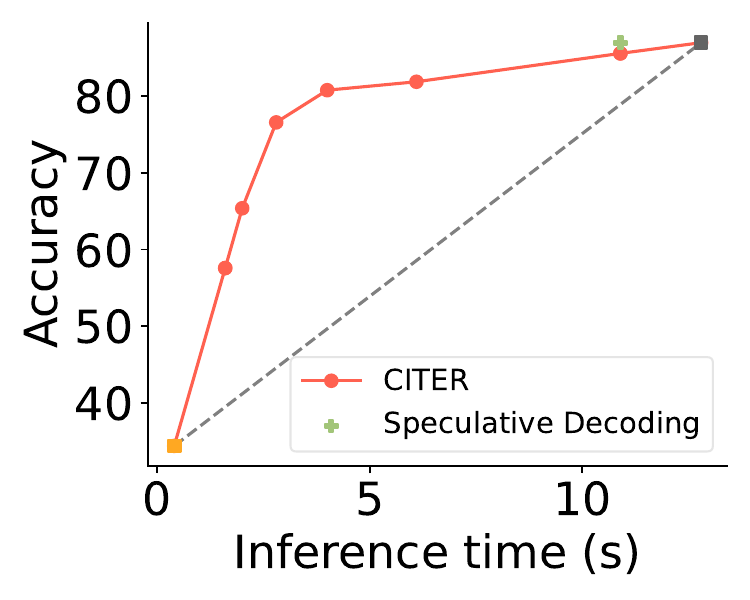}
   \caption{Accuracy vs. latency curve of \ours{} and Speculative Decoding.}
   \label{fig:latency}
   \vspace{-2em}
\end{wrapfigure}

\vspace{-1em}
\subsection{Analysis of the Impact of SLM Model Size}
\label{subsec:7b}
\vspace{-0.5em}

We further scale up the SLM size from Qwen2-1.5B to Qwen2-7B, while keeping the LLM fixed to Qwen2-72B, to understand the scalability of our framework. As shown in Figure~\ref{fig:7b}, the results clearly demonstrate that \ours{} reduces inference costs by up to 10\% while maintaining the same level of accuracy or achieves up to 11\% higher accuracy with the same cost when using Qwen2-7B as the SLM compared to Qwen2-1.5B, particularly on the commonsense QA and GSM8k datasets, underscoring our framework's scalability with larger SLMs. However, the performance gap is most noticeable when only very little tokens are generate by the LLM introducing a very small additional cost, and it gradually diminishes or even disappears as the cost further increases. This is expected, as the SLM's capacity limits its performance, and the quality of responses increasingly depends on the LLM as more calls are routed to it.

\vspace{-1em}
\subsection{Latency Analysis}
\label{subsec:latency_analysis}
\vspace{-0.5em}

In addition, we also evaluate the wall-clock latency of \ours{} against Speculative Decoding. The results are presented in Figure~\ref{fig:latency}. The plot shows that \ours{} offers a flexible trade-off between latency and accuracy. For instance, \ours{} achieves an accuracy of 80.8\% with a latency of only 4 seconds. While Speculative Decoding reaches a slightly higher accuracy of 87.0\%, it requires a significantly longer latency of 10.9 seconds. \ours{} can also achieve 85.6\% accuracy which is very similar to Speculative Decoding with the same latency. In addition, key advantage of \ours{} lies in the flexibility to operate at much lower latency points with only a minor compromise in accuracy, a feature not available with Speculative Decoding. This demonstrates the superior efficiency and adaptability of our framework.

\vspace{-1em}
\subsection{Qualitative Analysis on the Router}
\label{subsec:case}
\vspace{-0.5em}

\begin{figure*}
   \centering
   \includegraphics[width=0.9\linewidth]{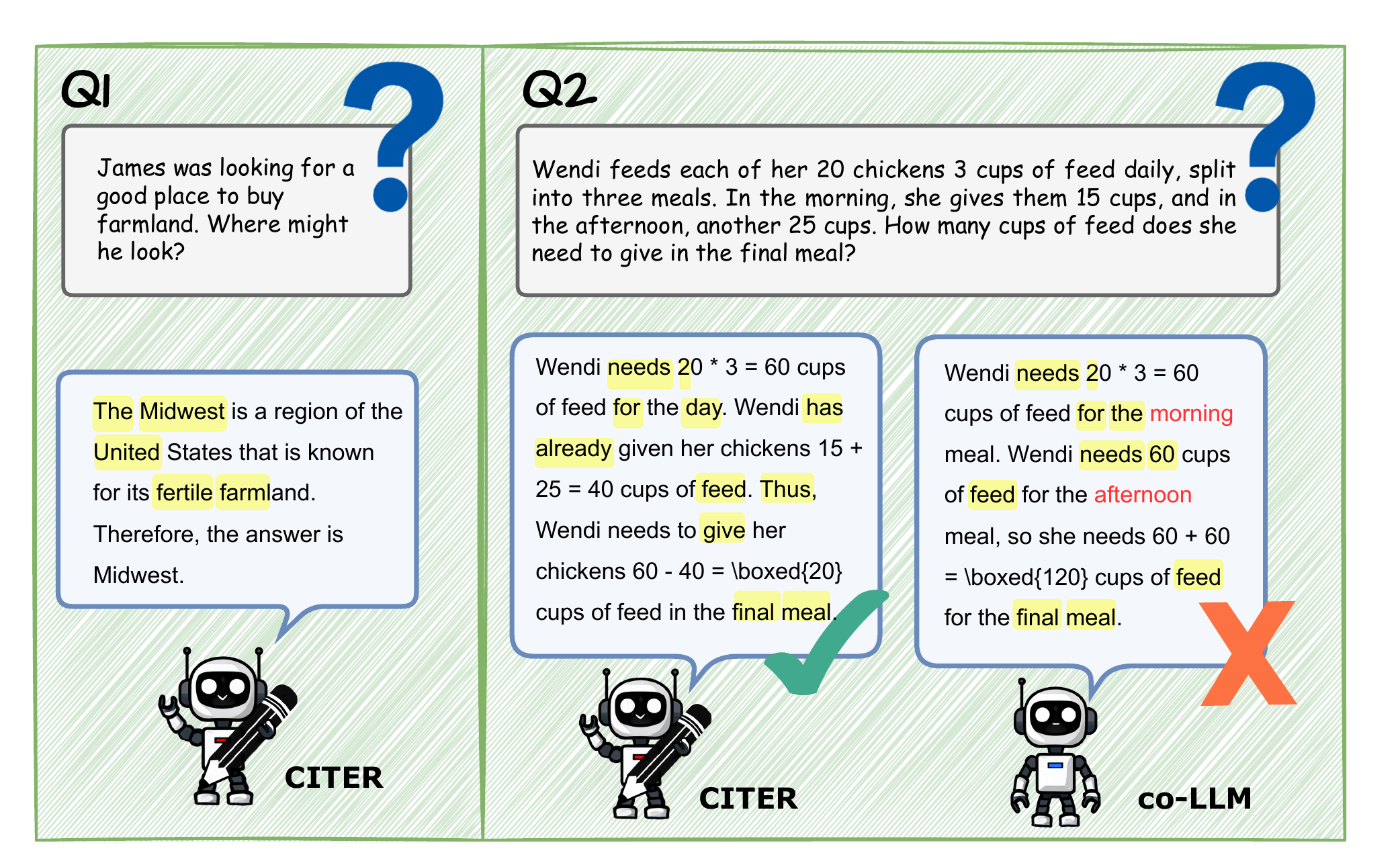}
   \caption{The case study analysis of our router. The words highlighted with yellow background are generated by the LLM, while other words are generated by the SLM. The red-marked words are the mistakes in Co-LLM's response.}
   \label{fig:case}
   \vspace{-1em}
\end{figure*}

Finally, we perform a case study to further analyze the decision-making process of the router in our framework. A selection of examples, along with their corresponding routing decisions, is shown in Figure~\ref{fig:case}. In the left example, it is clear that our router accurately identifies the critical tokens, including the first occurrence of the answer "Midwest" and the word "fertile," which describes the farmland in the Midwest, both crucial to the final answer. Moreover, most non-critical tokens are efficiently offloaded to the SLM, effectively reducing inference costs.

In the right example, we compare \ours{} with the token-level routing method Co-LLM. Clearly, our router outperforms Co-LLM by correctly identifying potential critical tokens, particularly time-related words. In Co-LLM's response, at the first red-marked word "morning," Co-LLM incorrectly routes the word "the" to the LLM while assigning the contextually important word "morning" to the SLM, leading to an initial error in the response. Similarly, Co-LLM routes the critical phrase "afternoon meal" to the SLM, resulting in the final incorrect prediction. In contrast, our router correctly identifies the critical word "day" and routes it to the LLM, followed by routing the phrase "has already" to accurately capture the reasoning process, ultimately leading to the correct prediction.

These examples illustrate that the router in \ours{} effectively distinguishes between critical and non-critical tokens, offloading non-critical tokens to the SLM to minimize inference costs, while leveraging the LLM to ensure the quality of the generated response.

\section{Related Work}
\label{sec:related}

In this section, we conduct a literature review that mainly focuses on prior LLM inference acceleration methods, especially those that involve using routing mechanisms and collaborative inference between LLMs for inference acceleration.
\paragraph{Query-Level Routing Mechanisms.} Previous routing methods~\citep{jang2023exploring,chronopoulou2023adaptersoupweightaveragingimprove,diao2023mixtureofdomainadaptersdecouplinginjectingdomain,lu2023routing,cheng2024damdynamicadaptermerging,lu2024twinmergingdynamicintegrationmodular,chen2023frugalgpt,wang2024fusing,srivatsa-etal-2024-harnessing,stripelis-etal-2024-tensoropera} for efficient inference mainly focus on routing entire user queries to different models for generation. For example, Routoo~\citep{mohammadshahi2024routoolearningroutelarge} proposes a performance predictor and a cost-aware decoder to route between LLMs, considering both performance and resource constraints; Hybird LLM~\citep{ding2024hybridllmcostefficientqualityaware} proposes a probabilistic router to select LLM backend for each query; RouteLLM~\citep{ong2024routellm} formulates the routing problem as a classification problem and employs a data augmentation framework to significantly expand the dataset used for training the router; \cite{gupta2024language} first leverages the small model (SLM) to generate the entire sequence, and then the decision to defer to the large model (LLM) is made based on the uncertainty (e.g., entropy or confidence derived from logits) during generation;FrugalGPT~\cite{chen2023frugalgpt} formulates the routing problem as a constrained optimization problem, where the final generated quality is maximized under a budget or inference cost constraint.
However, as highlighted in Section~\ref{sec:intro}, routing at the query-level granularity may lead to suboptimal performance, as non-critical tokens in complex queries may be generated inefficiently, while critical tokens in simple queries may suffer from inaccuracy. In contrast, token-level routing methods offer more fine-grained control over the routing process, improving both inference costs and the quality of the generated response.

\paragraph{Token-Level Routing Mechanisms.} Unlike query-level routing methods, previous token-level routing methods~\citep{pfeiffer2021adapterfusion,belofsky2023tokenleveladaptationloraadapters,muqeeth2024learningroutespecializedexperts,wang2024loraflowdynamiclorafusion,wu2024mixtureloraexperts,xu2024meteoramultipletasksembeddedlora} mainly focus on routing input tokens to different specialized experts to enhance performance without considering the inference costs. For example, Arrow~\citep{ostapenko2024modularllmsbuildingreusing} builds a mixture-of-experts (MoE) architecture with multiple LoRAs, dynamically routing inputs to different LoRAs during inference. Similarly, Branch-Train-MiX~\citep{sukhbaatar2024branchtrainmixmixingexpertllms} fine-tunes LLMs on different domains from a seed LLM, creating specialized experts to form an MoE framework.
Besides these methods, \cite{narasimhan2024fastercascadesspeculativedecoding} introduces token-level cascading and tackles the challenge of implementing deferral rules by leveraging speculative decoding, relying exclusively on the logits output by the model to determine when to defer to the LLM. Similarly, Co-LLM~\citep{shen2024learning} introduces a router to route tokens to models of different sizes. However, they only consider the current outputs from SLM and LLM when generating ground truth labels to make the router decisions. This may lead to suboptimal performance since the influence of current decisions on future tokens is not considered. Moreover, similar to other token-level routing methods, they focus on enhanced response quality without taking the inference costs of the inference process into account. In contrast, our CITER framework considers both the current token and the future impact of each decision, enabling more accurate and efficient routing.
\paragraph{Other Methods for LLM Inference Acceleration.}
In addition to routing methods, several approaches ranging from algorithmic to system optimizations~\citep{miao2023specinfer,kwon2023efficient,chen2024magicdec} have been proposed to accelerate LLM inference. Speculative Decoding~\citep{leviathan2023fastinferencetransformersspeculative,chen2023accelerating} employs a small draft model to generate potential next tokens, which are concatenated with previously generated tokens. These guesses are then processed by the target LLM in parallel to verify their correctness. Tokens are only committed to the final output if confirmed by the target LLM. Although this approach reduces inference time by generating multiple tokens in a single forward pass, it does not lower the overall computational complexity (e.g., the total amount of FLOPs). Speculative Streaming~\citep{bhendawade2024speculativestreamingfastllm} addresses the computational overhead of Speculative Decoding by predicting n-grams instead of individual tokens in each forward pass. However, it requires redesigning the LLM architecture, necessitating re-pretraining, which is computationally prohibitive for many use cases. Medusa~\citep{cai2024medusa} mitigates the re-pretraining issue by adding auxiliary heads to the original LLM, allowing n-gram predictions without modifying the core model. These heads can be trained while keeping the original LLM frozen, thereby avoiding the need for re-pretraining. SpecInfer and Sequoia~\citep{miao2023specinfer,chen2024sequoia} leverage tree-based parallelism for decoding and verification to further accelerate inference.

\section{Conclusion}
\label{sec:conclusion}

In this paper, we introduced CITER, a novel collaborative inference with token-level routing framework designed to reduce the inference cost of LLM while maintaining high-quality generation. By dynamically routing non-critical tokens to a SLM and reserving the LLM for critical tokens, CITER achieves an efficient balance between inference cost and generation quality. We formulated the routing problem as a policy optimization task, where the router learns to make token-level decisions based on both immediate token quality and long-term impact. Furthermore, we introduced a shortcut for reward estimation to enhance training efficiency. Experimental results across five benchmark datasets demonstrate that CITER significantly reduces inference costs while preserving accuracy, offering a promising solution for real-time and resource-constrained applications.

\bibliography{colm2025_conference}
\bibliographystyle{colm2025_conference}

\appendix

\section{Acknowledgment}
This work was supported in part by the National Institutes of Health (NIH) under grants R01AG079291, R01AG085581 and R01HL165061, National Science Foundation (NSF) under grants CNS2414087, NSF BCS2040381, NSF IIS2123952, NSF IIS1955532, NSF IIS2123952, NSF IIS2311990, and the Semiconductor Research Corporation (SRC) AIHW award 2024AH3210.

\section{Dataset Description}

In this section, we describe our benchmark datasets with more details.

\subsection{Commonsense QA}

CommonsenseQA is a large-scale, multiple-choice question-answering dataset designed to challenge and evaluate systems on their ability to leverage commonsense knowledge. The dataset consists of 12,102 questions, each accompanied by one correct answer and four distractor (incorrect) options, requiring models to distinguish the correct answer by understanding various types of commonsense reasoning. What sets CommonsenseQA apart is its emphasis on requiring a broader array of everyday knowledge, involving not only basic facts but also causal, temporal, and conceptual reasoning.

\subsection{ARC-Challenge}

The AI2 ARC dataset is a comprehensive collection of 7,787 grade-school-level multiple-choice science questions, meticulously curated to stimulate advancements in question-answering systems. The dataset is strategically divided into two subsets: the ARC-Easy Set and the ARC-Challenge Set. The ARC-Challenge Set, which is the subset we utilized in our work, comprises a selection of particularly difficult questions. These questions were specifically included because they were misclassified by both a traditional retrieval-based algorithm and a word co-occurrence algorithm, making them a true test of a model’s ability to understand and reason through complex scientific concepts. The ARC-Challenge subset serves as an ideal benchmark for testing sophisticated models, as it presents questions that require more than surface-level understanding or simple pattern matching.

\subsection{MMLU-Professional Psychology}

The MMLU dataset is a comprehensive multitask benchmark that comprises multiple-choice questions across a vast range of knowledge domains, including subjects in the humanities, social sciences, hard sciences, and other fields. It covers 57 distinct tasks such as elementary mathematics, U.S. history, computer science, law, and more, aimed at evaluating a model’s general world knowledge and problem-solving capabilities.

In our work, we focused specifically on the “Professional Psychology” subset of MMLU. This subset contains questions rich in domain-specific terminology, including specialized terms related to psychology and, occasionally, biological concepts tied to psychological phenomena. It provides a robust test for assessing a model’s proficiency in understanding and reasoning within a specialized academic field, thus offering insights into the model’s capability to handle complex, domain-specific content.

\subsection{GSM8k}

GSM8k (Grade School Math 8k) is a dataset consisting of 8.5K high-quality, linguistically diverse grade school math word problems. Designed to evaluate and improve question-answering capabilities in basic mathematical problem-solving, this dataset emphasizes multi-step reasoning, requiring between 2 and 8 steps to arrive at the correct solution.

The problems involve a sequence of elementary calculations using basic arithmetic operations—addition, subtraction, multiplication, and division—along with some early Algebra concepts. However, the dataset ensures that all problems are approachable for a bright middle school student, avoiding the need for advanced mathematical tools like variable definitions in most cases.

One of the distinctive features of GSM8K is that the solutions are presented in natural language rather than purely in mathematical expressions. This design decision aligns with the dataset’s goal to illuminate the reasoning capabilities of large language models (LLMs), specifically how they simulate an “internal monologue” when reasoning through problems. The dataset’s natural language solutions provide a more interpretable and instructive resource for evaluating the logical progression of LLMs in real-world tasks.

\subsection{MATH}

The Mathematics Aptitude Test of Heuristics (MATH) dataset consists of an extensive set of 12,500 intricate mathematical problems curated from prestigious competitions, such as the AMC 10, AMC 12, and AIME \cite{hendrycksmath2021}. Each problem is provided alongside a fully worked-out solution, offering step-by-step reasoning that facilitates both answer derivation and explanation generation. Covering a broad spectrum of mathematical topics—including Prealgebra, Algebra, Number Theory, Counting and Probability, Geometry, Intermediate Algebra, and Precalculus—the dataset serves as a rigorous benchmark for mathematical reasoning.

To enable a structured evaluation of model capabilities, problems are stratified into five difficulty levels (1 to 5), ensuring a progressive challenge across varying levels of complexity. The solutions are typeset in \LaTeX{}, maintaining precision and coherence in mathematical notation. Notably, the MATH dataset emphasizes problems that extend beyond direct formula application, frequently requiring creative heuristics and nontrivial problem-solving strategies. This characteristic makes it particularly valuable for training and assessing models in advanced mathematical reasoning and structured solution generation.

\section{Prompt}

In this section, we illustrate the prompt we used for each dataset.

\subsection*{Multiple-Choice Question Prompt}

For multiple-choice question datasets, including the Commonsense QA dataset, the ARC-Challenge dataset, and the MMLU-Professional Psychology dataset, we leverage the following prompt to require LLMs to provide both an explanation and a final answer in a specific format.

\textbf{Example:}

\begin{lstlisting}
Question: $question
Choices:
A. $choice_A
B. $choice_B
C. $choice_C
D. $choice_D
E. $choice_E

Please format your response in the following way: [Explanation]. Therefore, the answer is answer (label). Ensure the final sentence includes the answer followed by the label in parentheses.
Answer and Reasoning: $response
\end{lstlisting}

\subsection{GSM8k Math Problem Prompt}

For math problems in the GSM8k dataset and the MATH, the following prompt is employed to require LLMs to provide a step-by-step solution with clear reasoning in LaTeX format, with the final answer enclosed in a box using $\backslash$boxed\{\}.

\textbf{Example:}
\begin{lstlisting}
Solve the following math problem step by step. Ensure all reasoning is shown clearly in LaTeX format, and the final answer should be enclosed using the LaTeX command \boxed{} to display it in a box.
Question: $question
Solution: $solution
\end{lstlisting}

\section{Examples}

In this section, we list examples of each dataset, including the questions, prompts, and responses from LLMs to clearly demonstrate the prompt we use and the task of each dataset.

\subsection{Commonsense QA}

\begin{lstlisting}
Question: What do people aim to do at work?
Choices:
A. complete job
B. learn from each other
C. kill animals
D. wear hats
E. talk to each other

Please format your response in the following way: [Explanation]. Therefore, the answer is answer (label). Ensure the final sentence includes the answer followed by the label in parentheses.
Answer and Reasoning: The answer should be something that people aim to do at work. People generally aim to complete jobs. Therefore, the answer is complete job (A).    
\end{lstlisting}

\subsection{ARC-Challenge}

\begin{lstlisting}
Question: Juan and LaKeisha roll a few objects down a ramp. They want to see which object rolls the farthest. What should they do so they can repeat their investigation?
Choices:
A. Put the objects in groups.
B. Change the height of the ramp.
C. Choose different objects to roll.
D. Record the details of the investigation.

Please format your response in the following way: [Explanation]. Therefore, the answer is answer (label). Ensure the final sentence includes the answer followed by the label in parentheses.
Answer and Reasoning: To repeat their investigation, Juan and LaKeisha should record the details of the investigation. This will involve noting down the objects they rolled, the height of the ramp, and the distance each object rolled. This way, they can replicate the exact conditions and compare the results of their repeated investigation. This is the crucial step in the scientific method, as it ensures that the results are reliable and can be trusted. Therefore, the final answer is record the details of the investigation (D).    
\end{lstlisting}

\subsection{MMLU-Professional Psychology}

\begin{lstlisting}
Question: Justice primarily involves ___________.
Choices:
A. Recognizing fairness for all persons
B. Promoting accuracy and truthfulness
C. Benefitting others and do no harm
D. Respecting cultural and individual difference

Please format your response in the following way: [Explanation]. Therefore, the answer is answer (label). Ensure the final sentence includes the answer followed by the label in parentheses.
Answer and Reasoning: Justice primarily involves recognizing fairness for all individuals. It is about ensuring that everyone is treated equally and fairly, without any discrimination or bias. Therefore,  the answer is Recognizing fairness for all persons (A).    
\end{lstlisting}

\subsection{GSM8K}

\begin{lstlisting}
Solve the following math problem step by step. Ensure all reasoning is shown clearly in LaTeX format, and the final answer should be enclosed using the LaTeX command \boxed{} to display it in a box.
Question: Janet's ducks lay 16 eggs per day. She eats three for breakfast every morning and bakes muffins for her friends every day with four. She sells the remainder at the farmers' market daily for \$2 per fresh duck egg. How much in dollars does she make every day at the farmers' market?
Solution: 16 - 3 = << 16 - 3 = 13 >> 13 eggs are left after breakfast. 13 - 4 = << 13 - 4 = 9 >> 9 eggs are left after baking. 9 * 2 = << 9 * 2 = 18 >> 18 dollars she makes at the farmers' market. The final answer is: \\boxed{18}.
\end{lstlisting}

\subsection{MATH}

\begin{lstlisting}
Solve the following math problem step by step. Ensure all reasoning is shown clearly in LaTeX format, and the final answer should be enclosed using the LaTeX command \boxed{} to display it in a box.
Question: How many vertical asymptotes does the graph of $y=\\frac{2}{x^2+x-6}$ have?
Solution: The vertical asymptotes of a rational function occur at the values of $x$ for which the denominator of the function is equal to zero. \n\nSo, to find the vertical asymptotes of the function $y=\\frac{2}{x^2+x-6}$, we need to solve the equation $x^2+x-6=0$ for $x$.\n\nThis is a quadratic equation, and we can solve it by factoring. The equation $x^2+x-6=0$ can be factored into $(x-2)(x+3)=0$.\n\nSetting each factor equal to zero gives the solutions $x=2$ and $x=-3$.\n\n Therefore, the graph of $y=\\frac{2}{x^2+x-6}$ has $2$ vertical asymptotes, at $x=2$ and $x=-3$.\n The final answer is: \\boxed{2}
\end{lstlisting}

\section{Data Transformation Calculation and the Memory-bound LLM Generation}
\label{app:costs}

In this appendix, we first analyze the computational characteristics of large language models (LLMs) during the decoding phase and then conclude that this phase is fundamentally memory-bound on modern hardware, such as NVIDIA H100 GPUs. Finally, we illustrate the calculation of the data transformation required for each token during the decoding process.

To analyze the generation process of LLMs, we begin by illustrate their structures. Each Transformer block in a decoder-only LLM consists of the following components:
\begin{itemize}[leftmargin=*]
   \item \textbf{LayerNorm}
   \item \textbf{Multi-Head Self-Attention (MHSA)}: includes linear projections for queries (Q), keys (K), and values (V), scaled dot-product attention, and an output projection.
   \item \textbf{Residual Connection}
   \item \textbf{LayerNorm after Self-Attention}
   \item \textbf{Feedforward Network (FFN)}: typically two linear layers with an activation in between, often of shape $d \rightarrow 4d \rightarrow d$.
   \item \textbf{Residual Connection}
\end{itemize}

During decoding, tokens are generated one at a time. To avoid recomputation of attention over previous tokens, modern implementations cache the key and value projections from previous steps in GPU memory, referred to as the \textit{KV cache}. Moreover, \textit{FlashAttention} is employed to efficiently compute attention within a single fused kernel, minimizing data movement and maximizing usage of on-chip memory.

Because of these optimizations, each new token only requires computing its query vector and performing attention against cached keys and values. This reduces both computation and data movement compared to training or prompt processing.

Let $d$ be the hidden dimension, $l$ the number of layers, $m$ the length of the current context (i.e., number of cached tokens), and assume float16 precision (2 bytes per element). We now analyze the compute and memory access for each component in a single Transformer layer during decoding of one token:

\begin{itemize}[leftmargin=*]
   \item \textbf{LayerNorm}: Requires reading and writing a $d$-dimensional vector, calculating the mean and variance and used them for nomalization. \\
         \textit{Memory:} read $d$ inputs, wriet $d$ outputs, $2d$ memory access in total. \\
         \textit{Compute:} $4d$ FLOPs.

   \item \textbf{Q projection}: Matrix-vector product ($1 \times d$ multiply $d \times d$). \\
         \textit{Memory:} read $d$ inputs and weights ($d^2$), write $d$ outputs, $d^2+2d$ memory access in total.\\
         \textit{Compute:} $2d^2$ FLOPs

   \item \textbf{K/V projection}: Not needed during decoding, as keys/values are cached.

   \item \textbf{Attention (FlashAttention)}:
         \begin{itemize}
            \item Read $m \cdot d$ cached keys and $m \cdot d$ cached values.
            \item Compute attention scores and weighted sum over $m$ past tokens.
         \end{itemize}
         \textit{Memory:} read $2md$ inptus, wriet $d$ outputs, $2md+d$ memory access in total.\\
         \textit{Compute:} $4md+2m$ FLOPs (QK matmul + attention weighted sum)

   \item \textbf{Output projection}: Matrix-vector product ($1 \times d$ multiply $d \times d$).\\
         \textit{Memory:} read $d$ inputs and weights ($d^2$), write $d$ outputs, $d^2+2d$ memory access in total.\\
         \textit{Compute:} $2d^2$ FLOPs

   \item \textbf{FFN}: Two linear layers: $d \rightarrow 4d \rightarrow d$ with an activation in between\\
         \textit{Memory:} read $d+4d+4d$ inputs and weights ($8d^2$), write $4d+4d+d$ outputs, $8d^2+18d$ memory access in total.\\
         \textit{Compute:} $8d^2 + 4d + 8d^2 = 16d^2 + 4d$ FLOPs
\end{itemize}

Summing over all components, the total computation and memory per layer per token is:

\begin{align}
   \text{FLOPs per layer} & = 4d + 2d^2 + 4md+2m + 2d^2 + 16d^2+4d \notag \\
                          & = 20d^2 + 4md + 8d + 2m \text{ FLOPs}
\end{align}

\begin{align}
   \label{appeq:memory_access_per_layer}
   \text{Memory access per layer} & = 2*(2d + d^2+2d + 2md+d + d^2+2d + 8d^2+18d) \notag \\
                                  & = 20d^2+4md+50d \text{ bytes}
\end{align}

Assume a typical setup with $d = 8192$, $m = 1024$ (context length), float16 (2 bytes). The per layer FLOPs will be $20d^2 + 4md + 8d + 2m \approx 1.28 \text{GFLOPs}$ and the memory accessed per layer will be $20d^2+4md+50d \approx 1.28 \text{GB}$. Thus, the compute-to-memory ratio is $\approx 1$ FLOPs per byte

On the other hand, an NVIDIA H100 GPU has Peak FP16 Tensor Core throughput: $\sim 2000$ TFLOPs/s and peak memory bandwidth: $\sim 3$ TB/s. Thus, the compute-to-memory ratio: $\sim 666.67$ FLOPs per byte.

The actual compute-to-memory ratio of decoding is much lower ($\sim 1$ FLOPs/byte) than what the H100 GPU hardware is capable of ($\sim 666.67$ FLOPs/byte). Therefore, decoding in LLMs is significantly \textbf{memory-bound}: performance is bottlenecked by memory bandwidth rather than compute throughput. This suggests that optimizations that reduce memory movement can have a substantial impact on inference speed and that the memory movement amount can be a great metric for the theoretical analysis of inference speed.

Therefore, in our experiments, we record the generated source of each token for both the query-level routing methods and the token-level routing methods and leverage Equation~\ref{appeq:memory_access_per_layer} to calculate the data transformation amount that occurred during the whole generation process. Similarly, for speculative decoding, the generated source of each token is also recorded, and a similar equation, where the output token of one forward pass is changed from a single token to multiple tokens, is employed to calculate the data transformation amount. All the additional data transformation introduced by additional structure (such as the router) in those methods are also included properly. In addition, we deploy the SLM and the LLM on the same device, so there is no switch cost and additional data transformation when we switch between those two models. Finally, we employ the data transformation amount to indicate the computation cost of the generation process. This improves the reproducibility of our experimental results and avoids the result deviation caused by different hardware devices and experimental environments due to the selection of indicators such as inference time.

\end{document}

%% file: math_commands.tex
\usepackage{amsmath,amsfonts,bm}

\def\eqref#1{equation~\ref{#1}}

\def\1{\bm{1}}

\DeclareMathAlphabet{\mathsfit}{\encodingdefault}{\sfdefault}{m}{sl}
\SetMathAlphabet{\mathsfit}{bold}{\encodingdefault}{\sfdefault}{bx}{n}

\DeclareMathOperator*{\argmax}{arg\,max}